%% file: root.tex
\documentclass[letterpaper, 10 pt, journal, twoside]{IEEEtran}  %

\usepackage{graphicx} %
\usepackage{hyperref}
\hypersetup{
    colorlinks=true,
}

\usepackage{verbatim}
\usepackage{booktabs}
\usepackage{multirow, makecell, stackengine}
\usepackage{color}

\renewcommand{\arraystretch}{1.3}
\newcommand\m[1]{%
\renewcommand\arraystretch{1}%
\begin{tabular}{@{}l@{}}#1\end{tabular}}

\newcommand{\rev}[1]{{#1}}

\usepackage{caption}
\usepackage{subcaption}
\usepackage{multirow}

\title{DSEC: A Stereo Event Camera Dataset for Driving Scenarios}

\author{Mathias Gehrig, Willem Aarents, Daniel Gehrig and Davide Scaramuzza%

\thanks{Manuscript received: October 15th, 2020; Revised January 19th, 2021; Accepted February 16th, 2021.}%
\thanks{This paper was recommended for publication by Editor Cesar Cadena Lerma upon evaluation of the Associate Editor and Reviewers' comments.
This work was supported by the Swiss National Science Foundation (SNSF) through the National Center of Competence in Research (NCCR) Robotics and the SNSF-ERC Starting Grant.
} %
\thanks{The authors are with the Robotics and Perception Group, Dep. of Informatics, University of Zurich, and Dep. of Neuroinformatics, University of Zurich and ETH Zurich, Switzerland (\protect\url{http://rpg.ifi.uzh.ch}). }
\thanks{Digital Object Identifier (DOI): see top of this page.}
}

\makeatletter
\let\@oldmaketitle\@maketitle%
\renewcommand{\@maketitle}{\@oldmaketitle%
\includegraphics[width=0.495\textwidth]{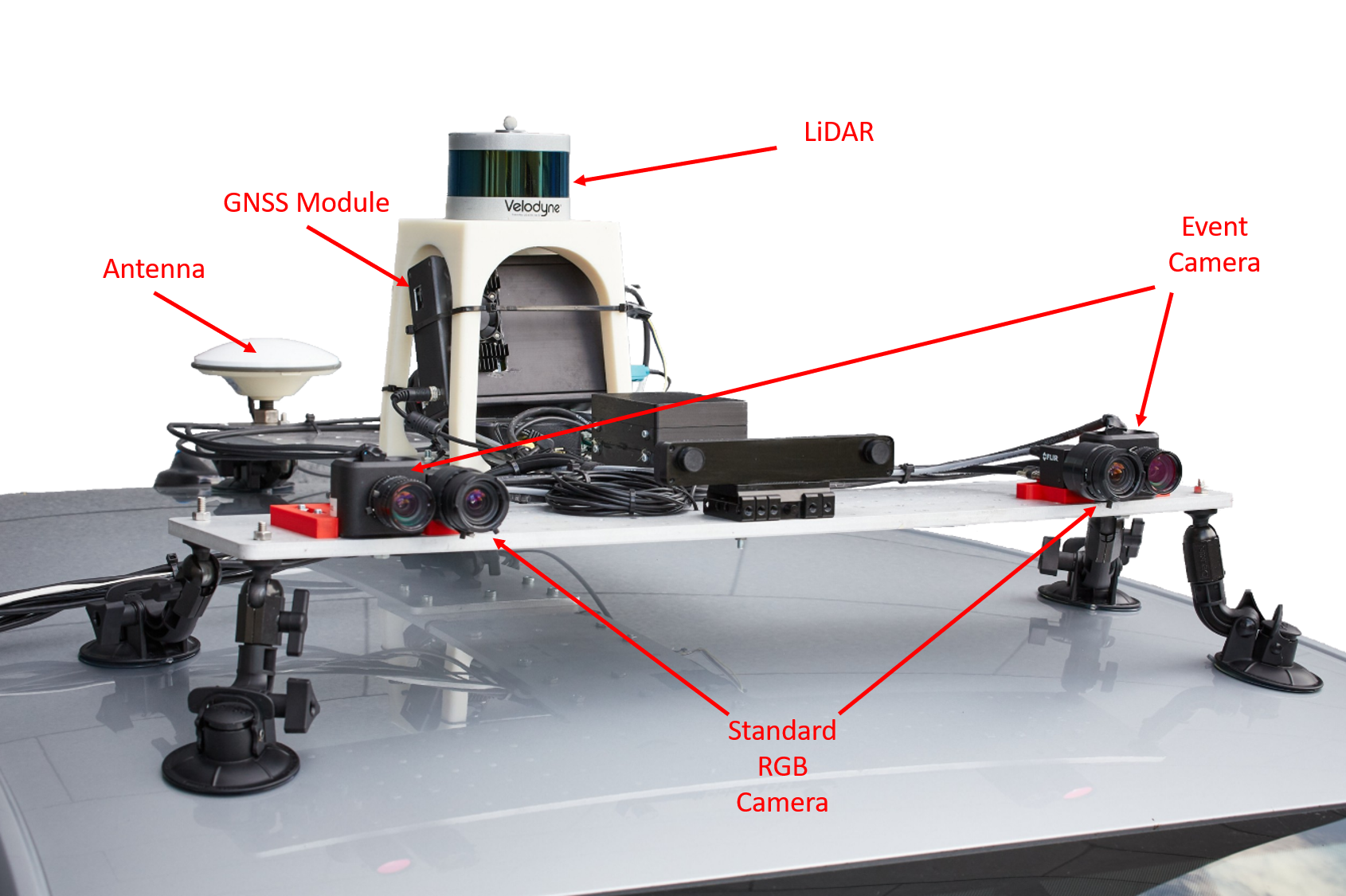}
\includegraphics[width=0.495\textwidth]{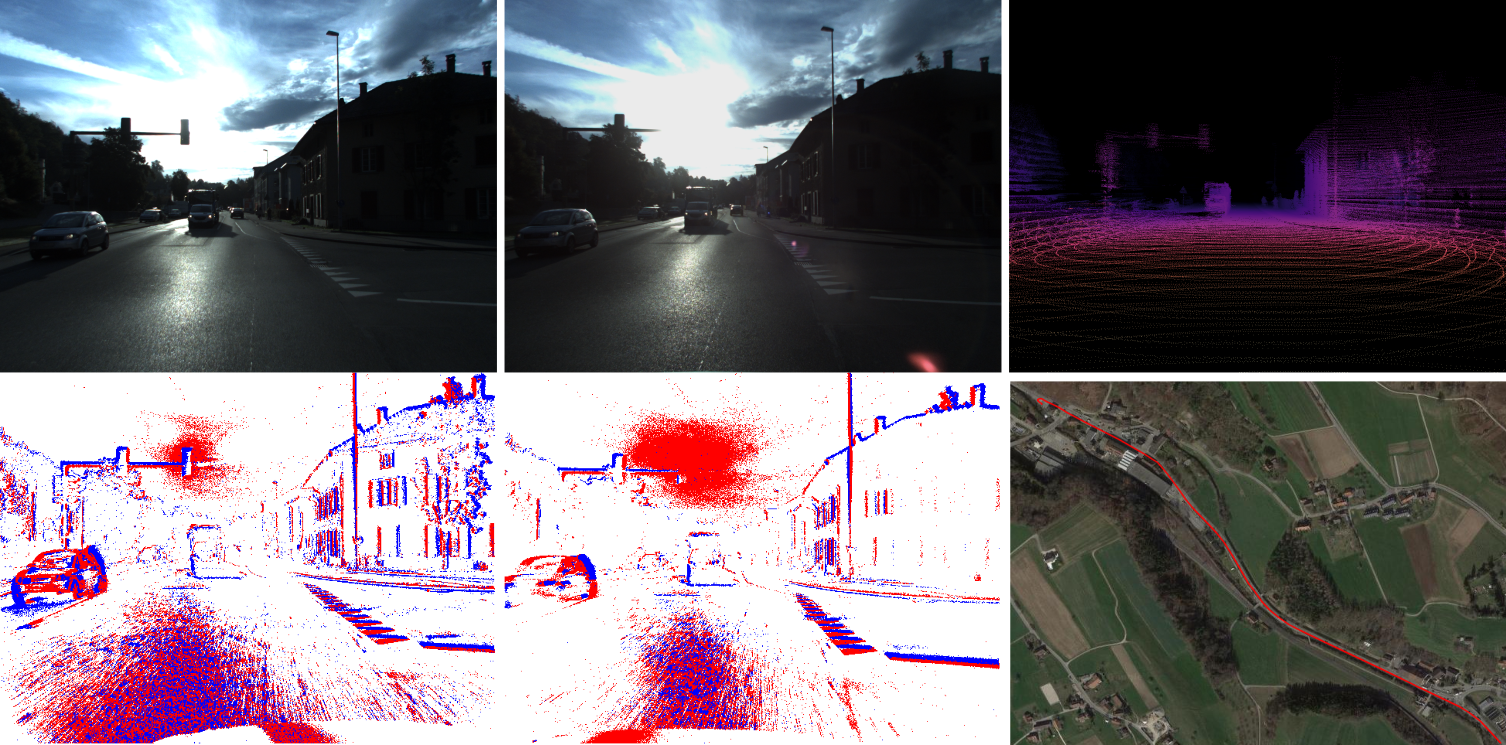}
\captionof{figure}{Left: The sensor suite mounted on top of a car, ready for data collection. Right:  Stereo RGB and event camera facing direct sunlight. The dataset contains depth maps for stereo matching and RTK GPS for accurate position measurement.}\label{fig:eyecatcher}
}%
\makeatother

\begin{document}

\maketitle

\input{sections/abstract}

Code and dataset: \url{http://rpg.ifi.uzh.ch/dsec.html}
\input{sections/introduction}
\input{sections/relatedwork}
\input{sections/dataset}

\input{sections/groundtruth}
\input{sections/experiments}

\section{Discussion}
During the recording sessions, we made several observations that might be of interest to users of this dataset.

\begin{figure*}
    \centering
    \includegraphics[width=0.32\textwidth]{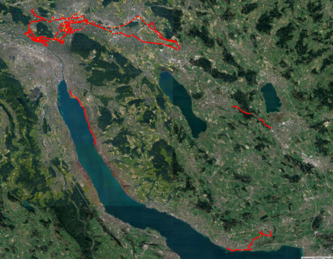}
    \includegraphics[width=0.32\textwidth]{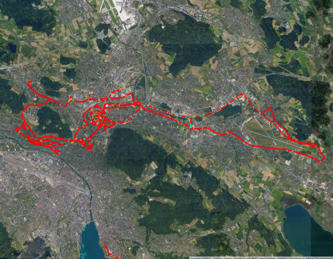}
    \includegraphics[width=0.32\textwidth]{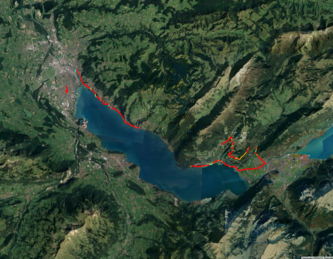}
    \caption{Overlay of GPS trajectories onto a satellite map: The trajectories are colored according to the quality of the GPS reception. Red, orange and cyan indicates highest, medium or lower reception quality respectively. Data was recorded in various urban, suburban and rural areas in Switzerland. The left image shows the region around Lake Zurich. The center image is the section showing the trajectories around the city of Zurich while the right image shows the region of Lake Thun.}
    \label{fig:gpstraj}
\end{figure*}

\begin{figure*}[h!]
    \centering
    \includegraphics[width=0.7\textwidth]{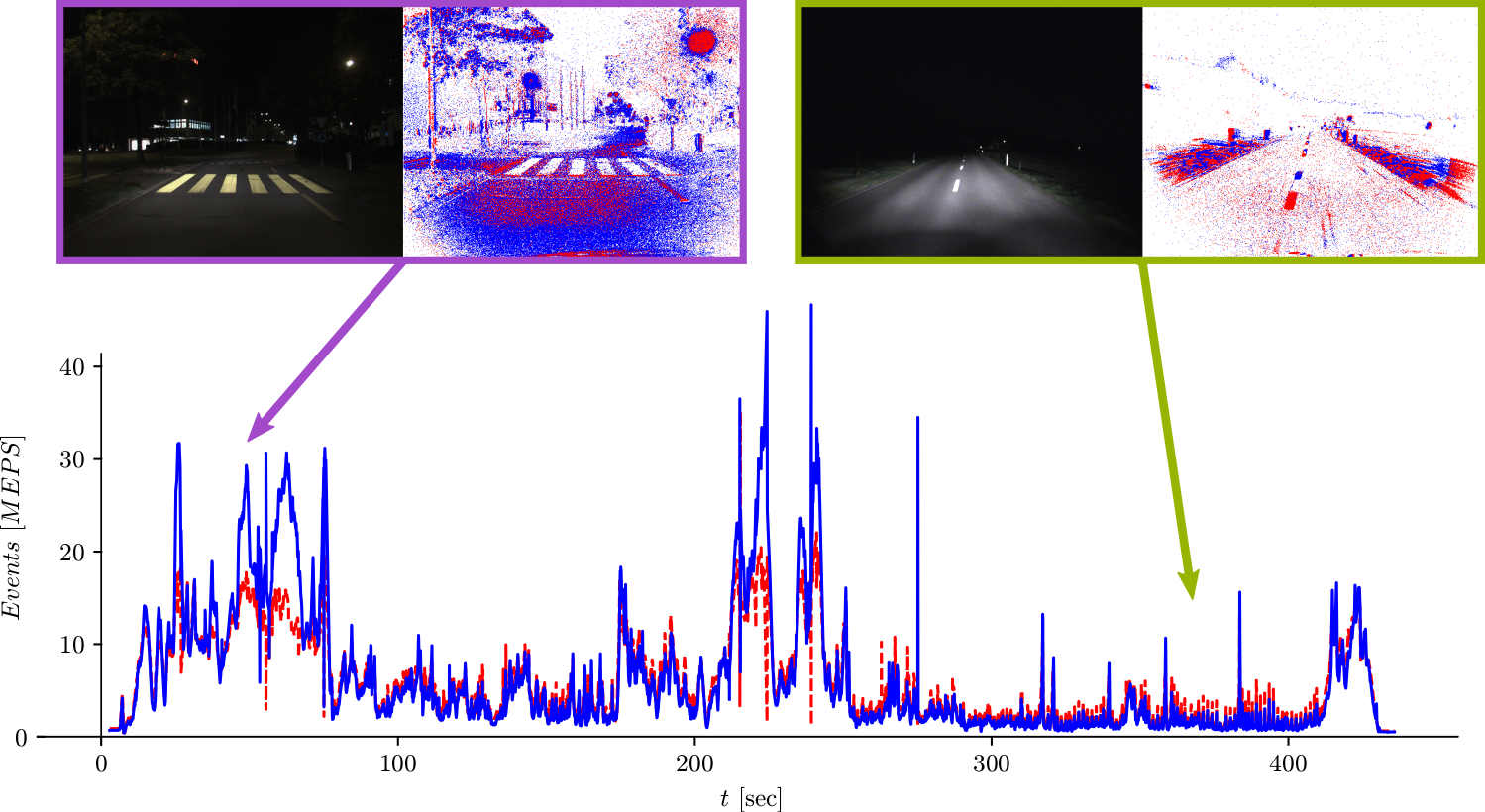}
    \caption{The event rate in millions per second (MEPS) \rev{during a night sequence}. The red and blue curves correspond to negative and positive event polarity respectively. The upper left pair of RGB and event frame corresponds to a situation where a street lamp flashes and periodically triggers events for a large number of pixels. In contrast, the upper right pair of RGB and event frame depicts a scene without artifical light sources other than the car. In this scenario, the event rate is much lower due to the absence of periodically flashing light sources.}
    \label{fig:night_event_rate}
\end{figure*}

Some of the provided sequences contain situations where the cameras face the sun directly, as shown in figure \ref{fig:eyecatcher} or \ref{fig:challenging}. In such scenarios, the contrast of \rev{the cameras} is significantly reduced due to lens flare. We tuned the biases and settings of the event cameras to the best of our knowledge. Still, we believe that higher quality lenses and better lens coating could reduce the lens flare effect in the future.

On the other hand, we also provide sequences during night time. Night driving is challenging for standard and event cameras. Standard cameras, for example, rely on increasing the exposure time, which introduces a significant amount of motion blur. The event camera increasingly produces noisy events due to the reduced signal-to-noise ratio of the signal. An often overlooked difficulty for event cameras at night are flashing lights. Figure \ref{fig:night_event_rate} highlights this effect and its influence on the spiking event rate while the cameras are in close proximity to a flashing street lamp. Rarely, the event readout of the event cameras could saturate and slightly delay the events.

\section{Limitations}
\rev{The groundtruth depth generation approach in section \ref{sec:depth_gt}, while handling occlusions and moving objects, introduces biases into the disparity groundtruth. First, the filtering approach leads to increased sparsity of disparity labels on moving objects. Second, the filtering approach is based on SGM and performed on the images. Hence, the resulting disparities are sparser in challenging conditions such as night driving or oversaturation of the images.}

\section{Conclusion}
This work proposes DSEC, the first large-scale event camera dataset in driving scenarios with a wide-baseline and high resolution event cameras. We additionally record RGB frames that are time-synchronized to the event cameras to foster research in sensor fusion. We believe that this dataset serves as a foundation to further explore the utility of event cameras in safety-critical applications such as autonomous driving. In the future, we plan to continuously extend the dataset with new tasks, improved hardware, and corner cases.

\section*{Acknowledgment}
The authors would like to thank Mario Millh{\"a}usler for his contributions to the hardware design.

\IEEEtriggeratref{28}
\bibliographystyle{ieeetr}
\bibliography{all}

\end{document}

%% file: sections/abstract.tex
\begin{abstract}

Once an academic venture, autonomous driving has received unparalleled corporate funding in the last decade. Still, operating conditions of current autonomous cars are mostly restricted to ideal scenarios. This means that driving in challenging illumination conditions such as night, sunrise, and sunset remains an open problem. In these cases, standard cameras are being pushed to their limits in terms of low light and high dynamic range performance. To address these challenges, we propose, DSEC, a new dataset that contains such demanding illumination conditions and provides a rich set of sensory data. DSEC offers data from a wide-baseline stereo setup of two color frame cameras and two high-resolution monochrome event cameras. In addition, we collect lidar data and RTK GPS measurements, both hardware synchronized with all camera data. One of the distinctive features of this dataset is the inclusion of high-resolution event cameras. Event cameras have received increasing attention for their high temporal resolution and high dynamic range performance. However, due to their novelty, event camera datasets in driving scenarios are rare. This work presents the first high resolution, large scale stereo dataset with event cameras. The dataset contains 53 sequences collected by driving in a variety of illumination conditions and provides ground truth disparity for the development and evaluation of event-based stereo algorithms.

\end{abstract}

%% file: sections/introduction.tex
\section{Introduction}

\IEEEPARstart{A}{utonomous} driving has received increasing attention in the robotics and computer vision communities over the past decade. However, the robust performance of vision systems in challenging conditions is still an open problem. 

Event cameras show promise to alleviate these challenges by providing complementary sensory information to standard cameras. In contrast to standard cameras that capture frames at regular intervals, event cameras report per-pixel brightness changes as a stream of asynchronous events. Event cameras offer significant advantages compared to standard cameras, such as high temporal resolution (microseconds), high dynamic range ($>120$dB), less motion blur, and sub-millisecond latency, all of which are crucial for autonomous driving and automotive scenarios. For a detailed survey about event cameras, we refer to \cite{Gallego20pami}. 

While event camera research has gained momentum in the last few years, event camera datasets in driving scenarios are still a rarity. This can be attributed to the fact that the sensor has only recently been commercialized and thus does not yet enjoy the same availability of standard cameras. The shortage of large scale datasets impedes the event camera research community because it restricts the evaluation to small scale environments that can quickly saturate the performance or generalizability of proposed algorithms.

This work specifically addresses the lack of large scale stereo datasets. To the best of our knowledge, MVSEC \cite{Zhu18ral} is the only dataset with a stereo event camera in driving scenarios. A significant drawback of MVSEC for stereo algorithms is the small resolution of the DAVIS346 cameras ($0.1$ Megapixels) and the small baseline of 10 centimeters on top of that. This is in stark contrast to state-of-the-art standard benchmarks such as KITTI \cite{Geiger12cvpr, Menze2015CVPR, xie2016semantic} or DrivingStereo \cite{yang2019drivingstereo} with camera baselines of approximately 55 centimeters and camera resolution over 1 Megapixel (MP). Both aspects are crucial for stereo algorithms as it is well known that the depth error is approximately inversely proportional to the product of the baseline and focal length \cite{gallup2008variable}.

While event cameras certainly possess unique assets compared to standard cameras, estimating dense depth using only stereo event cameras is an ill-posed problem. In a static scene with constant illumination, a static event camera will not trigger any events and can therefore not perform stereo matching. In the case of a dynamic scene and a static camera, it is at least possible to estimate the depth of the dynamic part of the scene. Still, the majority of event-based stereo methods only estimate sparse or semi-dense depth \cite{zhou2018semi, schraml2015event, Zhu18eccv,  Andreopoulos18cvpr}. \rev{We refer to the survey by Steffen et al. for an overview of neuromorphic stereo vision approaches \cite{steffen2019neuromorphic}}. For this reason, we also equip our data collection setup with two high-resolution standard cameras that can allow for dense depth estimation even in cases of static scenes with no motion.

The fusion of frames and events has proven to be beneficial in several prior works \cite{Gehrig19ijcv, jiang2020learning, Scheerlinck19ral, Pan19cvpr, pan2020single}. Nevertheless, the underlying assumption of all these approaches is that frames and events share the same pixels. This is the case for example for the older DAVIS cameras \cite{Brandli14ssc} but not for state-of-the-art event cameras \cite{finateu20205}. Our data collection setup, depicted in figure \ref{fig:eyecatcher}, separates standard and event cameras spatially such that each stereo camera pair is mounted next to each other. By doing so, we minimize the parallax effect between event and standard cameras and facilitate sensor fusion of both modalities. Hence, the proposed dataset raises novel challenges for sensor fusion algorithms which can take advantage of high resolution event and standard cameras.

Finally, we collect data in both favorable and very challenging illumination conditions such a depicted in fig. \ref{fig:challenging}. For real-world applicability, computer vision systems must show robustness also in the most adverse conditions such as driving at night, facing direct sunlight or fast moving objects. This dataset incorporates multiple sequences with precarious illumination conditions that challenge standard and event cameras. We believe that this dataset serves as a foundation to better understand the performance of novel algorithms in various illumination conditions and challenging situations.

In summary, we propose a new dataset that
\begin{itemize}
    \item addresses the need for a large scale outdoor stereo event camera datasets
    \item offers the novel challenge of combining events and frames without the common assumption of accurate pixel correspondence.
    \item provides sequences with various illumination conditions, from favorable to very challenging ones.
\end{itemize}

\begin{figure}
    \centering
    \addtocounter{figure}{-1}%
    \includegraphics[width=0.23\textwidth]{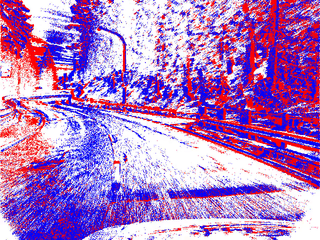}
    \includegraphics[width=0.23\textwidth]{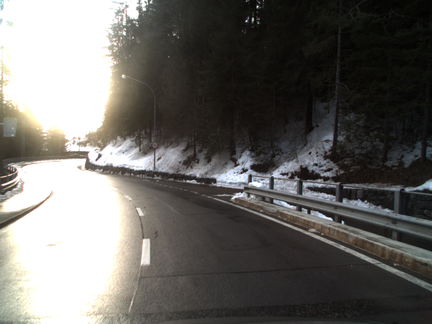}\\
    \includegraphics[width=0.23\textwidth]{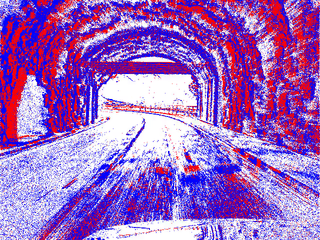}
    \includegraphics[width=0.23\textwidth]{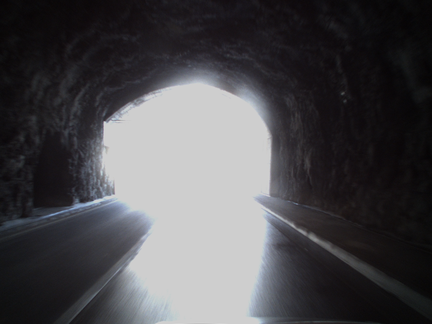}\\
    \includegraphics[width=0.23\textwidth]{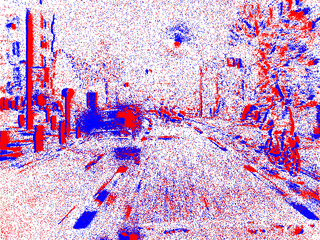}
    \includegraphics[width=0.23\textwidth]{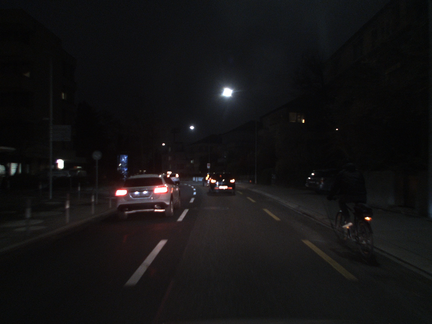}\\
    \caption{\rev{Examples of challenging illumination conditions from the recordings. The first two examples show that the frame-based camera is prone to saturation of pixels due to the lower dynamic range. The third row is a driving scene at night in which the bicyclist on the right side of the street is difficult to spot.}}
    \label{fig:challenging}
\end{figure}

%% file: sections/relatedwork.tex
\section{Related Work}

A multitude of driving datasets for computer tasks has been released in the last decade. This section serves as a brief overview of datasets in the driving domain.

Following the seminal work of Geiger et al. \cite{Geiger12cvpr}, multiple datasets have been published mostly for more specialized purposes. 
KITTI 2015 \cite{Menze2015CVPR} extended the original KITTI version to provide scene flow ground truth. Later KITTI 360 \cite{xie2016semantic} proposed a novel labeling scheme and data for 2D and 3D semantic segmentation. Similarly, Cityscapes \cite{Cordts2016Cityscapes} is a widely used benchmark for image-based semantic segmentation. 
Since then, a number of large scale datasets for semantic segmentation have been released \cite{neuhold2017mapillary, huang2018apolloscape, geyer2020a2d2, yu2020bdd100k}. 
The Waymo open dataset \cite{sun2020scalability} and nuScenes \cite{caesar2020nuscenes} instead focus mainly on object detection or tracking. Oxford RobotCar \cite{Maddern16ijrr} and KAIST Urban \cite{jeong2019complex} target localization and mapping, while DDAD \cite{guizilini20203d} and DrivingStereo \cite{yang2019drivingstereo} are specialized for monocular or stereo depth estimation respectively. KAIST MultiSpectral differs in that their datasets contains data from a thermal camera. Instead of capturing real-world data, Synthia \cite{ros2016synthia} and \cite{Gaidon16cvpr} provide perfect ground truth in synthetic environments. However, all aforementioned datasets do not contain event camera data.

\input{snippets/relwork_table}

In this paragraph, we outline datasets in the automotive domain that contain event camera data. Table \ref{tab:datasets} provides an overview.
The ADD dataset \cite{de2020large} contains a single low resolution (304x240) event camera for 2D object detection. This dataset was recently extended with a high resolution (1 MP) event camera \cite{perot2020learning} for the same task. N-CARS \cite{Sironi18cvpr} contains very short (100 ms) sequences for binary classification of cars. DET \cite{cheng2019det} is a dataset  captured with a CeleX-V DVS camera for driving lane extraction. Brisbane-Event-VPR \cite{fischer2020event} presents a driving dataset with a color DAVIS346 camera for place recognition. DDD17 \cite{binas2017ddd17} and DDD20 \cite{hu2020ddd20} are datasets comprising many hours of driving data from a monochrome DAVIS346 camera featuring various vehicle control data. These event camera datasets do not contain stereo event cameras and can thus not be used for the development of stereo matching algorithms.

Closest to our work is the MVSEC dataset \cite{Zhu18ral}. It contains two DAVIS346B cameras with a resolution of roughly 0.1 MP and a baseline of 10 cm. Furthermore, the dataset features a stereo monochrome frame-based camera setup with a similar baseline and a resolution of 0.4 MP. What our dataset has in common with MVSEC is the LIDAR sensor. Both utilize a Velodyne VLP16 for the computation of ground truth depth data in post-processing. Different from MVSEC, however, we provide event camera data with 3 times higher resolution from two Prophesee Gen 3.1 sensors with a baseline of 60 cm similar to KITTI 360 \cite{xie2016semantic}. In addition to that, we mount two RGB global shutter cameras each next to the event cameras with a baseline of 51 cm. The combination of significantly higher camera resolution and larger baseline therefore addresses one of the main limitations of the driving sequences of the MVSEC dataset.

%% file: snippets/relwork_table.tex
\begin{table*}[]
\centering
\begin{tabular}{llllllll}
\toprule
\multirow{2}{*}{\textbf{Dataset}} & \multicolumn{3}{l|}{\textbf{Frame Camera}}                                               & \multicolumn{2}{l}{\textbf{Event Camera}} & \multirow{2}{*}{\textbf{Lidar}} & \multirow{2}{*}{\textbf{Groundtruth}} \\ \cline{2-6}
                                  & \multicolumn{1}{c}{Resolution {[}MP{]}} & Baseline {[}cm{]} & \multicolumn{1}{l|}{Color} & Resolution {[}MP{]}  & Baseline {[}cm{]}  &                                 &                                       \\ \midrule
ADD \cite{de2020large}                               &                                         &                   &                            & 0.1                  & -                  & -                               & 2D Bounding Box                       \\
1MP Detection \cite{perot2020learning}                     & 4                                       & -                 & yes                        & 0.9                  & -                  & -                               & 2D Bounding Box                       \\
N-CARS  \cite{Sironi18cvpr}                            & \multicolumn{3}{c}{}                                                                     & 0.01                 & -                  & -                               & Binary Class                          \\
DDD17 \cite{binas2017ddd17}                             & 0.1                                     & -                 & no                         & 0.1                  & -                  & -                               & Vehicle Control, GPS                  \\
DDD20 \cite{hu2020ddd20}                             & 0.1                                     & -                 & no                         & 0.1                  & -                  & -                               & Vehicle Control, GPS                  \\
DET  \cite{cheng2019det}                               & \multicolumn{3}{c}{}                                                                     & 1.0                  & -                  & -                               & Lane Extraction                       \\
Brisbane-Event-VPR  \cite{fischer2020event}                & 2.1                                     & -                 & yes                        & 0.1                  & -                  & -                               & GPS, Place Matches                    \\
MVSEC  \cite{Zhu18ral}                             & 0.4                                     & 10                & no                         & 0.1                  & 10                 & VLP-16                          & Depth, GPS                            \\ \midrule
\textbf{DSEC} (Ours)              & 1.6                                     & \textbf{51}       & yes                        & 0.3                  & \textbf{60}        & VLP-16                          & Depth, RTK GPS                        \\ \bottomrule
\end{tabular}

\caption{Comparison of different event camera datasets in driving scenarios. Empty or missing entries indicate \emph{not available} or \emph{not applicable} respectively. ADD, N-CARS and DET do not feature a frame camera. MVSEC and this work are the only dataset with a stereo camera setup. Unlike other event camera datasets, our dataset provides a sufficient baseline and camera resolution for stereo matching in driving scenarios. Note that a baseline of above 50 centimeters is consistent with the KITTI datasets \cite{Geiger12cvpr, xie2016semantic} and DrivingStereo \cite{yang2019drivingstereo}}
\label{tab:datasets}
\end{table*}

%% file: sections/dataset.tex
\section{Dataset}
\subsection{Sensors}

\begin{table}[h]
    \centering
    \begin{tabular}{ll}\toprule
        Sensor Type & Description\\\midrule
        $2\times$ Prophesee Gen3.1 & \m{Prophesee PPS3MVCD event camera\\$640\times 480$ pixels, $3/4$" CMOS
 \\Monochrome\\$\geq 120$ dB dynamic range}\\\midrule
        $2\times$ FLIR Blackfly S USB3& \m{FLIR BFS-U3-16S2C-CS\\$1440\times 1080$ pixels, $1/2.9$" CMOS\\ Color\\  $71.4$ dB dynamic range \\20 Hz frame rate}\\\midrule
        Velodyne VLP-16 & \m{16 Channels\\ 100m range \\$\pm3$cm range accuracy \\ 10 Hz rotation rate}\\\midrule
        U-Blox RTK GNSS Receiver & \m{ZED-F9P GNSS Receiver\\ RTK signal with 10 Hz update rate\\ $\pm1$ cm position accuracy}\\\bottomrule
    \end{tabular}
    \caption{Details about sensors utilized for data collection}
    \label{tab:sensors}
\end{table}
Our sensor suite consists of a multi-camera setup equipped with a Velodyne lidar and an RTK GPS system. We refer to figure \ref{fig:eyecatcher} for visualization and table \ref{tab:sensors} for details of the characteristics of each sensor.

The monochrome stereo event cameras have a resolution of $640\times 480$ and are mounted with a baseline of $60$ centimeters. Next to each event camera, a color FLIR Blackfly S USB3, with a resolution of $1440\times 1080$, is mounted in a forward-facing direction. The baseline of $51$ centimeters of the frame cameras is slightly smaller than the event camera baseline to account for the higher resolution of the Blackfly camera. We deliberately minimized the baseline between the two camera types on each side of the stereo setup. By doing so, we achieved a baseline of approximately $4.5$ centimeters on each side, which introduces only a small parallax and can be exploited for sensor fusion approaches.

The Blackfly cameras operate in auto exposure mode while prioritizing exposure time adjustments over increasing the analog gain to reduce noise. The maximum exposure time of $15$ ms is typically only reached at night time, in which case the gain increases to minimize motion blur. The frames are captured in 12-bit raw format at 20 Hz. We apply a high-quality demosaicing algorithm \cite{menon2006demosaicing} before mapping the debayered image to 8 bits.

The parameter settings, usually called biases, for event cameras differ fundamentally from standard cameras. Event cameras do not have an exposure setting. Instead, the sensitivity of the camera can be adjusted to be more responsive to changes in contrast. In bright daylight, we manually set the sensitivity to a higher value which means that the contrast threshold is lower. As a consequence, more events are generated and more detail in the scene is revealed. This is especially important in direct sunlight where lens flare reduces the contrast. On the other hand, we increase the contrast threshold in dawn/dusk and night conditions to reduce the number of noisy events.

\subsection{Sequences}
\begin{table}[ht!]
\begin{tabular}{@{}lllll@{}}
\toprule
\textbf{Split} & \textbf{Area} & \textbf{Time} & \textbf{\#Sequences} & \textbf{Duration {[}s{]}} \\ \midrule
\textbf{Train} & Interlaken    & Day           & 5                    & 367                       \\
               & Thun          & Day           & 1                    & 12                        \\
               & Zurich City   & Day, Night    & 35                   & 2255                      \\ \cmidrule(l){4-5} 
               &               &               & \textbf{41}          & \textbf{2634}             \\ \midrule
\textbf{Test}  & Interlaken    & Day           & 3                    & 251                       \\
               & Thun          & Day           & 2                    & 64                        \\
               & Zurich City   & Day, Night    & 7                    & 244                       \\ \cmidrule(l){4-5} 
               &               &               & \textbf{12}          & \textbf{559}              \\ \midrule
\textbf{Total} &               &               & \textbf{53}          & \textbf{3193}             \\ \bottomrule
\end{tabular}
\caption{Details about the dataset sequences. In total, the dataset comprises close to an hour of driving data in diverse illumination conditions while driving in urban and rural environments.}
\label{tab:dataset_details}
\end{table}

Figure \ref{fig:gpstraj} illustrates the trajectories of the data recording. Over 4 TB raw data was recorded in urban, suburban, and rural areas in Switzerland. From the raw recordings we carefully select sequences with good recording and calibration quality. As summarized in table \ref{tab:dataset_details}, the final dataset consists of 53 sequences in three different areas of Switzerland at day and night time.

\begin{figure*}[t!]
  \centering
    \includegraphics[width=0.7\textwidth]{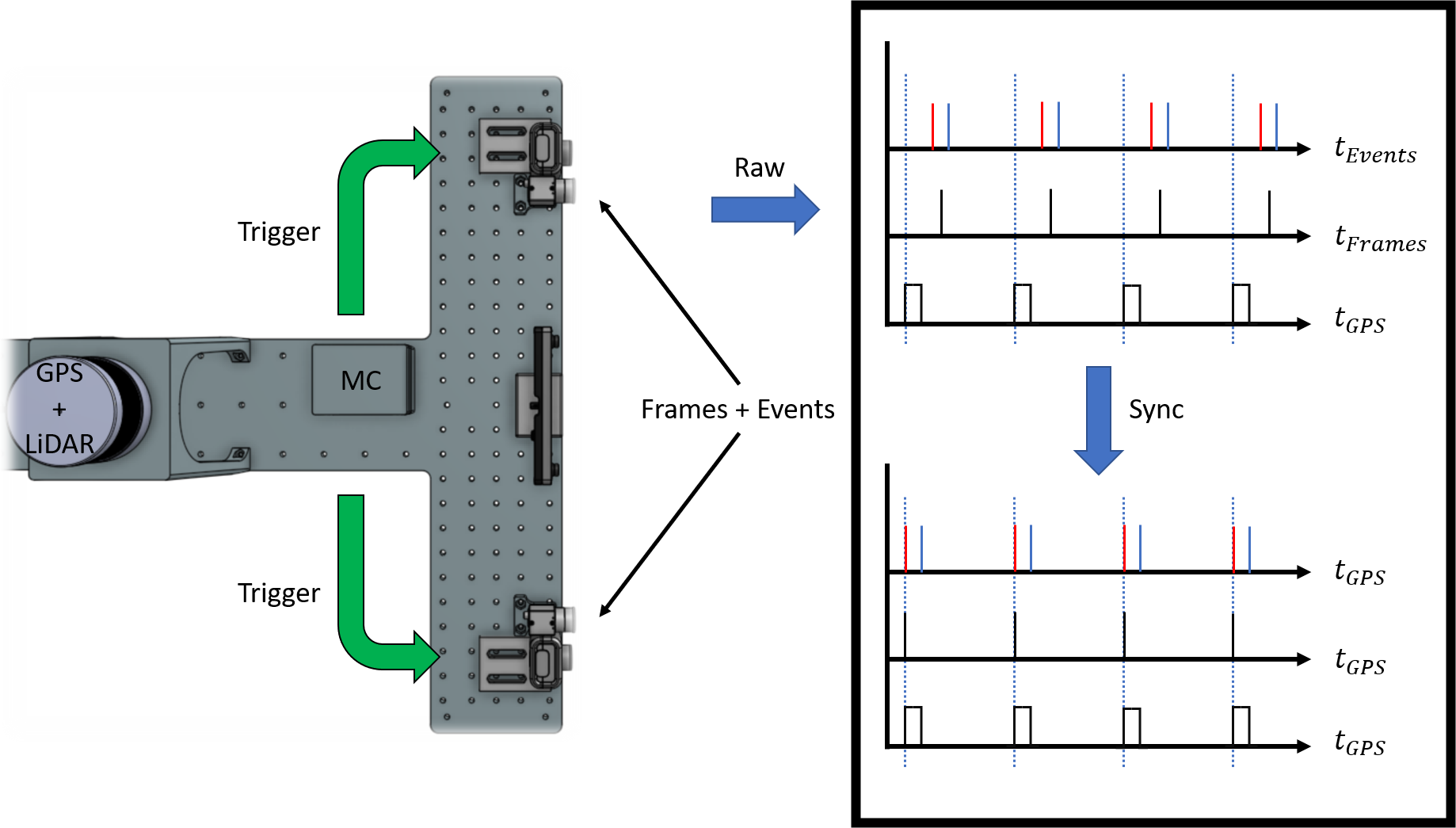}
  \caption{The sensor suite used to record DSEC. It includes a GPS receiver (GPS), a Velodyne VLP16 (LIDAR), a microcontroller (MC), two Prophesee Gen3.1M Event Cameras (VGA resolution), and two FLIR BlackFly S USB3 RGB cameras. The microcontroller is synchronized to the GPS module and sends a trigger signal at 20 Hz to both event cameras and both standard cameras. The event cameras record these pulses in the form of special events, which are precisely timestamped and thus allow synchronization with the microcontroller clock. The standard camera starts the exposure time whenever a trigger signal arrives.}
  \label{fig:setup_sync}
\end{figure*}

\section{Synchronization and Calibration}
This section describes the methodology for time synchronization of the sensors as well as the extrinsic and intrinsic calibration procedures. Time synchronization is achieved on hardware by using a microcontroller that associates GPS trigger and time signals with trigger signals sent to the cameras. The camera intrinsic and extrinsic calibration procedure uses E2VID, a neural network-based image reconstruction \cite{Rebecq19pami}, in combination with Kalibr \cite{Furgale13iros}. Finally, camera to lidar calibration is performed by manual rotational adjustment based on an accurate CAD model.
\subsection{Time Synchronization}
For an overview of the time synchronization procedure, see Fig.~\ref{fig:setup_sync}.
Time synchronization of the sensor suite was achieved through the use of hardware triggers supplied by the microcontroller (Fig.~\ref{fig:setup_sync} MC) and GPS receiver.
The GPS receiver first sends triggers in the form of \rev{PPS (pulse per second)} signals at the top of each second to the microcontroller and the lidar, which update their internal timestamp accordingly. 
This means that lidar measurements are already provided in GPS timestamps.
Upon receiving a PPS signal, the microcontroller forwards a sequence of 20 pulses to the events cameras and standard cameras.
Each time the event camera receives a pulse it records a so-called ``special event" which is precisely timestamped with the clock of the event camera. 
When a pulse arrives at the standard camera, it triggers the camera to start exposing. 
Thus the standard camera is triggered at 20 Hz, and the event cameras receive a stream of special events at 20 Hz. 
For synchronization, we use the timestamps provided by the GPS triggers to compute the pulse times as they are transmitted by the microcontroller.
Finally, we use the timestamps of the pulses to express frame timestamps, and event timestamps in the GPS clock.

\subsection{Camera Calibration}
\begin{figure}
    \centering
    \includegraphics[width=0.23\textwidth]{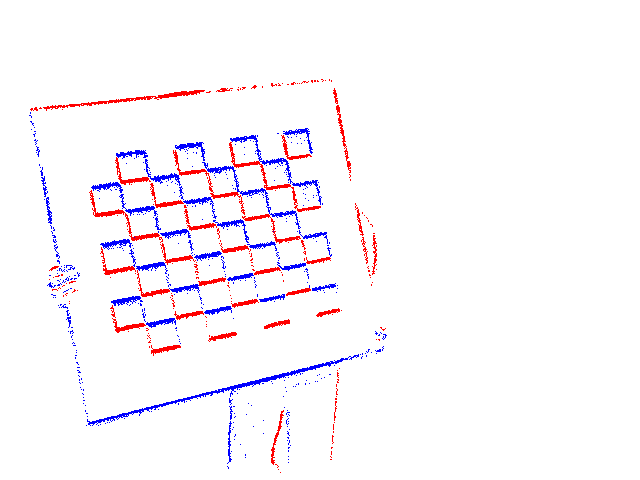}
    \includegraphics[width=0.23\textwidth]{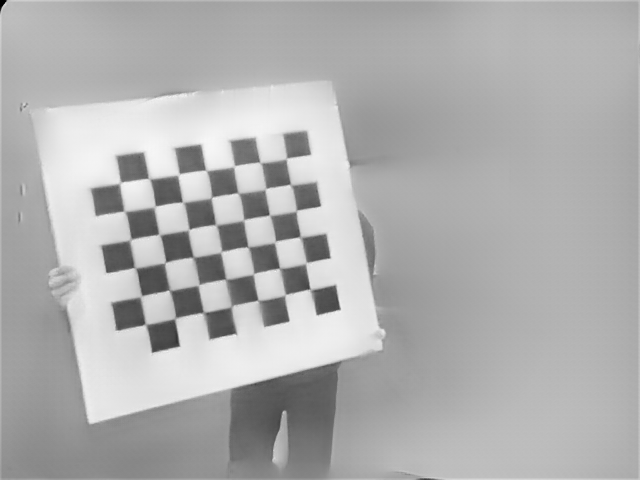}
    \caption{Events (left) are used to reconstruct images (right) at mid-exposure times of the frames. These reconstructions can then be used in conjuction with standard calibration frameworks.}
    \label{fig:img_rec}
\end{figure}

For the estimation of camera intrinsics and extrinsics, we use a checkerboard pattern together with the Kalibr toolbox \cite{Furgale13iros}. While this procedure is straightforward with standard cameras, it is not directly applicable to event cameras. Instead of relying on blinking patterns, we approach the problem with image reconstruction \cite{Rebecq19pami} from the event streams. Figure \ref{fig:img_rec} shows an example of reconstructed images from events that are directly used by Kalibr to detect corners and extract the checkerboard pattern. For accurate extrinsic calibration, it is important that all frames must be captured or reconstructed simultaneously. Hence, we can directly make use of our hardware time synchronization scheme to reconstruct images at the middle of the exposure times of the frames.

\subsection{Camera to Lidar Calibration}
\rev{The CAD model of the recording setup provides an initial estimate of the transformations between the different sensors. To further refine the initial estimate, we employ refinement steps using ICP. We 
select $N$ different views in a sequence of interest. For each pose,
\begin{itemize}
    \item compute a local pointcloud from the Lidar measurements,
    \item compute the 3D pointcloud resulting from running the SGM \cite{hirschmuller2007stereo} algorithm on the rectified image pair,
    \item align the stereo and Lidar pointcloud with a modified version of point-to-plane ICP \cite{chen1992object} that only optimizes for rotation. We do not require ICP to optimize for translation because the translation can be accurately retrieved from the CAD model.
\end{itemize}
Finally, we average the $N$ rotation quaternions \cite{markley2007averaging} for a more robust extrinsic calibration. $N$ is typically chosen between 2 and 5 but more poses can be used with this approach.
}

%% file: sections/groundtruth.tex
\begin{figure}[t!]
  \centering
    \includegraphics[width=0.48\textwidth]{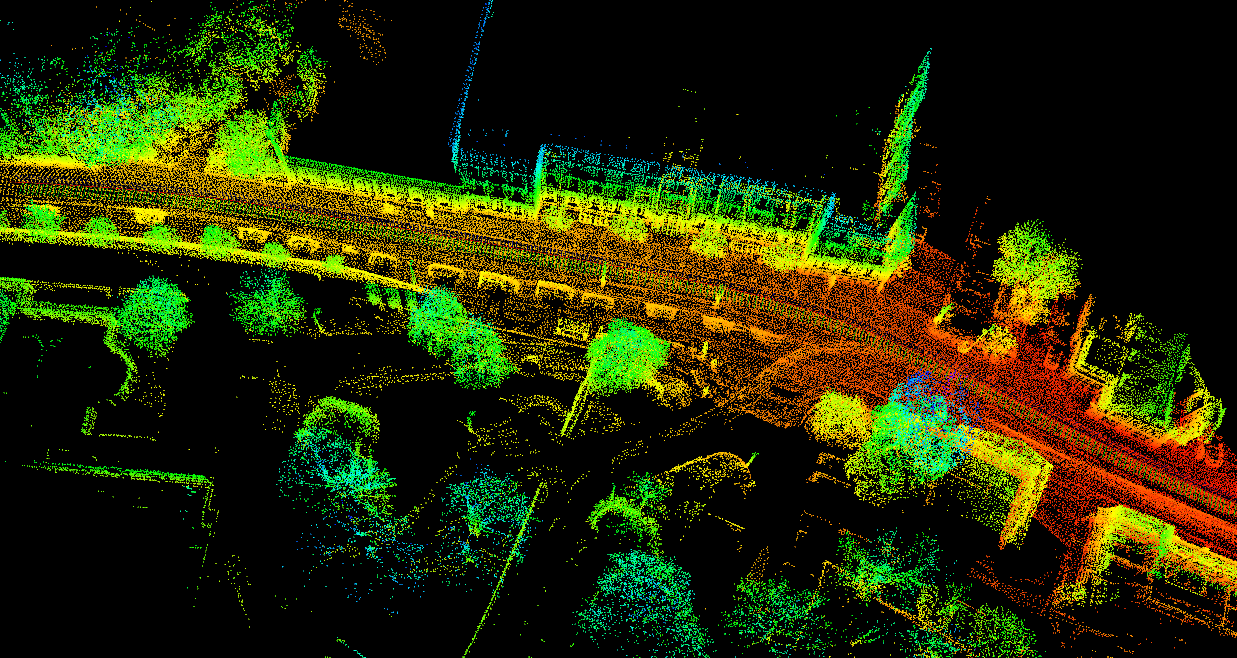}
  \caption{A local pointcloud that is used to generate depth ground truth via projection into camera views.}
  \label{fig:pointcloud}
\end{figure}

\section{Ground Truth Depth Generation}\label{sec:depth_gt}
\begin{figure*}
    \centering
    \includegraphics[width=0.24\textwidth]{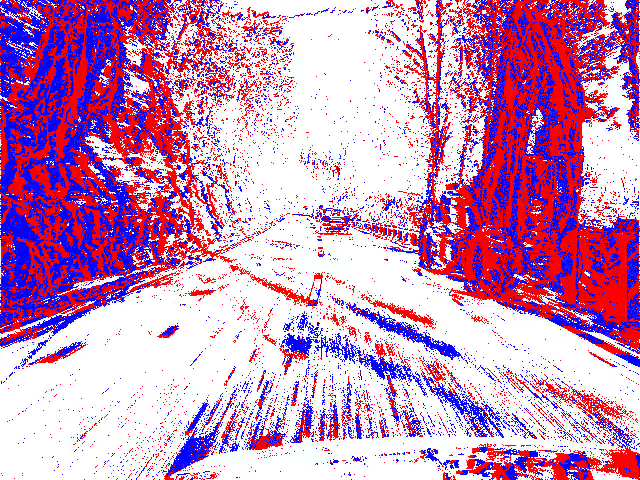}
    \includegraphics[width=0.24\textwidth]{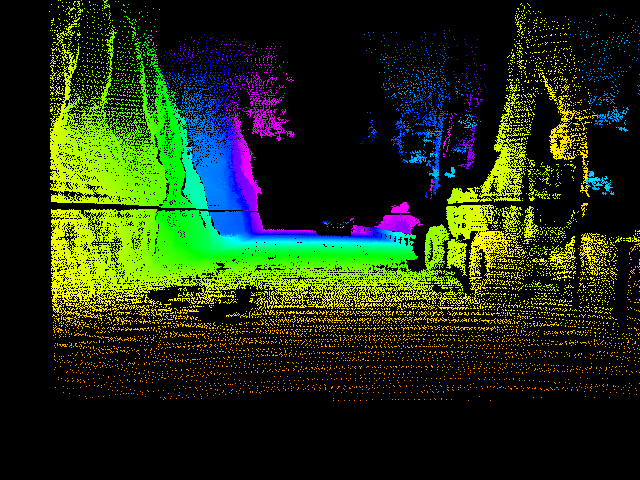}
    \includegraphics[width=0.24\textwidth]{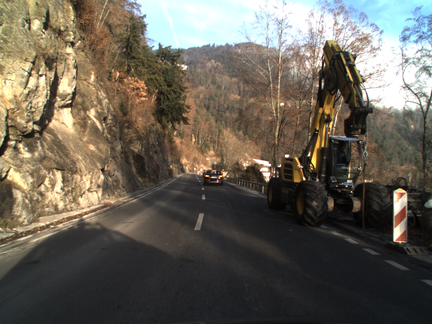}
    \includegraphics[width=0.24\textwidth]{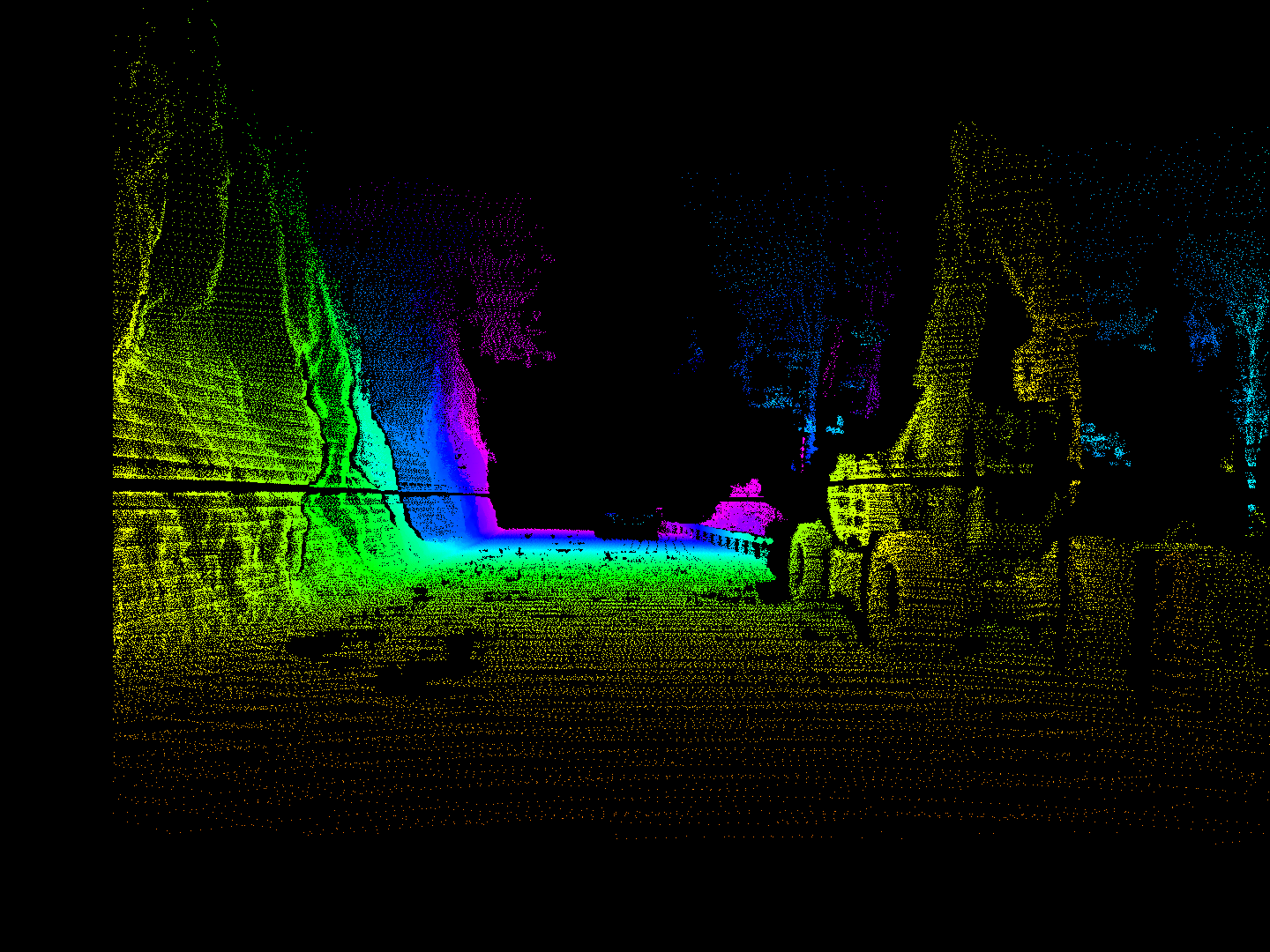}\\
    \includegraphics[width=0.24\textwidth]{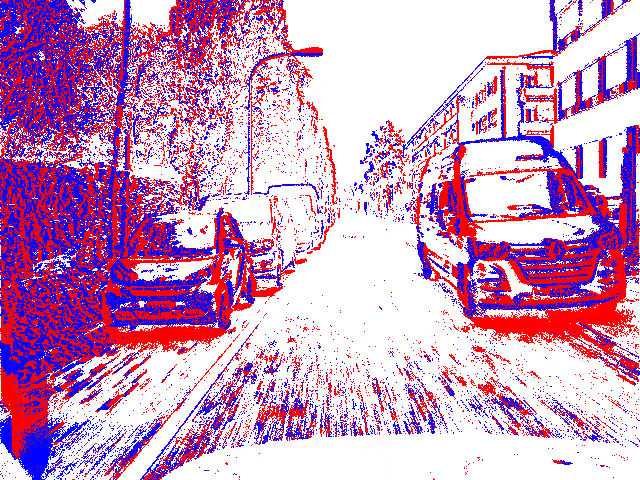}
    \includegraphics[width=0.24\textwidth]{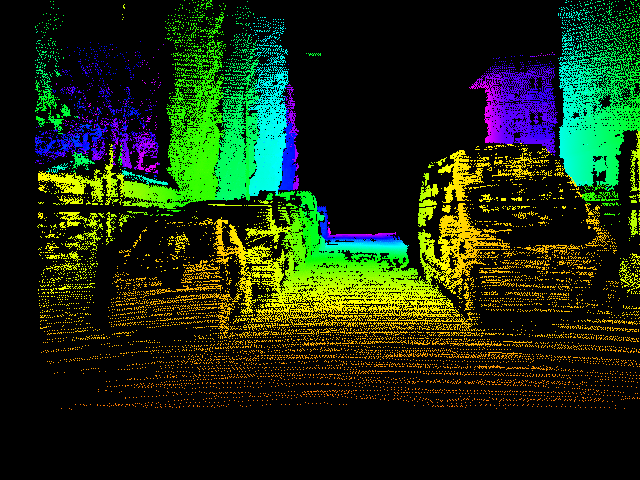}
    \includegraphics[width=0.24\textwidth]{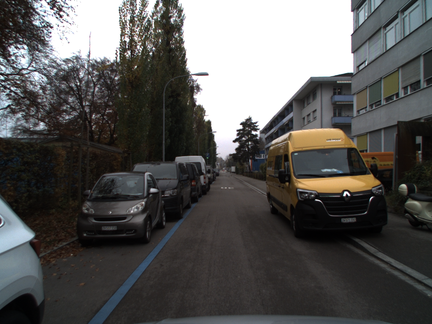}
    \includegraphics[width=0.24\textwidth]{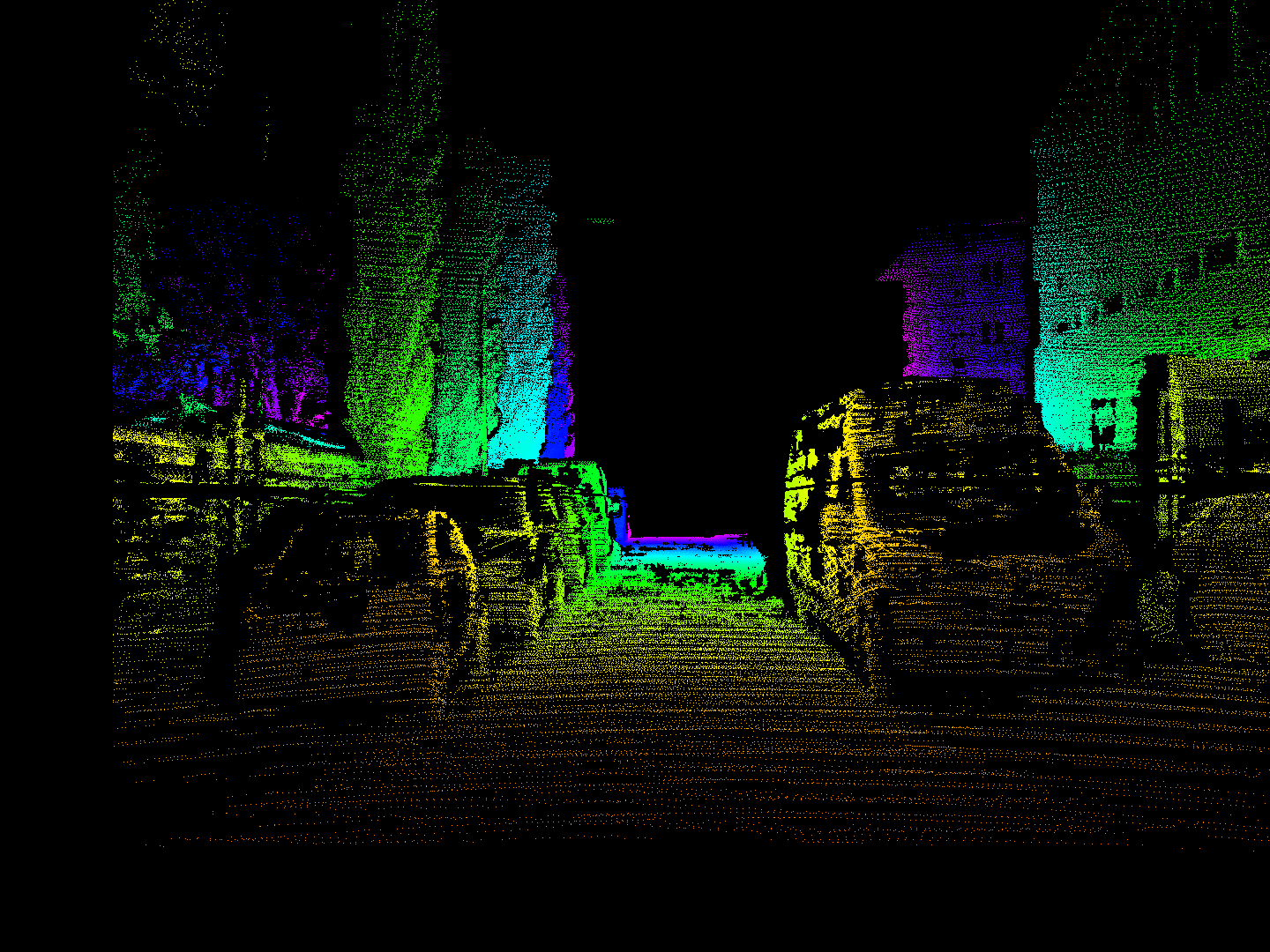}\\
    \caption{Examples of recorded event and image data with corresponding disparity groundtruth. From left to right: events with the corresponding disparity and image with corresponding disparity. The disparity image is colored to provide a sense of depth.}
    \label{fig:data_examples}
\end{figure*}

The previously obtained intrinsic and extrinsic calibration can now be used to generate groundtruth depth maps. A single lidar scan only provides very sparse depth information. To improve the ground truth density, LIO-Mapping \cite{ye2019tightly} computes the pose for each scan of the lidar. We query each pose and extract the relative poses of several preceding and proceeding scans. An example of such a pointcloud is shown in figure \ref{fig:pointcloud}.
\rev{Simply projecting this pointcloud into the camera frames is problematic because potential occlusions and  moving objects are not taken into account. To address this issue we implement a filtering method similar to those employed by the KITTI depth \cite{uhrig2017sparsity} and DrivingStereo dataset \cite{yang2019drivingstereo}:
\begin{itemize}
    \item Accumulate a local Lidar pointcloud for each view 
    \item Generate a disparity map from the rectified image pair using SGM \cite{hirschmuller2007stereo} which is not affected by moving objects.
    \item The lidar pointcloud is projected into the camera frame. Only the 3D points are retained that correspond to a disparity value close enough to the disparity at that pixel estimated by SGM. We remove 3D points further away than 50 meters to maintain an accurately filtered map.
    \item The filtered Lidar pointcloud is subsequently projected into the left event and frame cameras to generate the groundtruth disparity maps at 10 Hz.
\end{itemize}
}
Figure \ref{fig:data_examples} visualizes the resulting groundtruth disparity maps without artefacts from moving objects or occluded points. As an example of filtered groundtruth due to moving objects is the top row in figure \ref{fig:data_examples}. Even though the front car is approximately 40 meters from the camera, the filtering approach successfully removes erroneous points.

%% file: sections/experiments.tex
\section{Evaluation Metrics}\label{sec:eval_metric}
We evaluate disparity estimation methods using a number of established metrics in stereo matching. Following \cite{Geiger12cvpr, scharstein2014high}, we evaluate methods on this dataset with the $N$ pixel disparity error metric (D$N$). This metric computes the percentage of disparity predictions with error larger than $N$ pixels. We choose $N$ as 2 and 1 to assess the predictions in both coarse and fine-grained manner. Furthermore, as in the Middlebury dataset, we also evaluate the mean absolute disparity error (MAE) and root-mean-square disparity error (RMSE) both in pixels.

\section{Experiments}
In order to assess the difficulty of this dataset, the stereo method presented in \cite{Tulyakov_2019_ICCV} is evaluated on the test set. We use the checkpoint provided by the authors and feed 0.5 seconds of events into the event queue. The resulting high D$1$ error of 85.9 percent and MAE of 13.0 suggest that the model does not generalize to this dataset. Consequently, we retrain their model on the training set for 200k iterations without data augmentation and choose the best performing for the final evaluation. Note that our baseline implementation uses the voxel grid representation \cite{Zhu19cvpr} that contains events within a duration of 50 milliseconds. We empirically found that the performance is comparable to using event queues while significantly faster to train.

Table \ref{tab:baseline_experiment} provides a summary of the results of the baseline experiment. We compute the errors for each area individually and also distinguish day and night sequences in the Zurich area. To assess the overall performance, we compute the mean of the error of all predictions. Evidently, the error metrics are comparable between the different locations. However, the D$1$ error is approximately 50\% higher during night compared to the day sequences. This is to be expected because the signal to noise ratio decreases in low illumination conditions.
Overall, the evaluation suggests that stereo methods trained on MVSEC do not generalize to this dataset and must be retrained to reach acceptable performance. Furthermore, we implemented and evaluated a baseline method on which future work can improve.

\begin{table}[]
\centering
\begin{tabular}{@{}ll|llll@{}}
\toprule
Area                    & Time     & D1             & D2            & MAE            & RMSE           \\ \midrule
\textit{Interlaken}     & Day      & 10.57          & 3.09          & 0.568          & 1.357          \\
\textit{Thun}           & Day      & 10.84          & 3.13          & 0.621          & 1.595          \\
\textit{Zurich City}    & Day      & 10.50          & 2.35          & 0.550          & 1.289          \\
                        & Night    & 15.47          & 4.26          & 0.671          & 1.649          \\
                        & Total    & 11.28          & 2.65          & 0.568          & 1.345          \\ \midrule
\multicolumn{2}{l}{\textbf{Total}} & \textbf{10.91} & \textbf{2.90} & \textbf{0.574} & \textbf{1.379} \\ \bottomrule
\end{tabular}
\caption{\rev{Baseline evaluation on the test set. We evaluate the method presented in \cite{Tulyakov_2019_ICCV} on a total of 12 sequences in 3 different areas within Switzerland. Here we report the average scores separated by city and day or night time as well as the total score. D\textit{N} refers to the percentage of disparity errors higher than \textit{N} pixels. MAE is mean absolute error and RMSE is the root-mean-square error.}} \label{tab:baseline_experiment}
\end{table}